\documentclass[10pt,twocolumn,letterpaper]{article}

\usepackage[utf8]{inputenc}
\usepackage{times}
\usepackage{epsfig}
\usepackage{graphicx}
\usepackage{amsmath}
\usepackage{amssymb}
\usepackage{bm}
\usepackage{algorithm}
\usepackage{algpseudocode}
\usepackage{stmaryrd}
\usepackage{url}
\usepackage{hyperref}
\usepackage{multirow}

\usepackage{afterpage}

\newcommand\blankpage{%
    \null
    \thispagestyle{empty}%
    \addtocounter{page}{-1}%
    \newpage}

\usepackage{draftwatermark}
\SetWatermarkText{Draft}
\SetWatermarkScale{2}

\begin{document}

\title{Event-based Gesture Recognition with Dynamic Background Suppression using Smartphone Computational Capabilities}

\author{Jean-Matthieu Maro\\
Sorbonne Universités\\
Paris, France\\
{\tt\small corr@jmatthi.eu}
\and
Ryad Benosman\\
University of Pittsburgh\\
Carnegie Mellon University\\
Sorbonne Universités\\
{\tt\small benosman@pitt.edu}
}

\twocolumn[
  \begin{@twocolumnfalse}
    \maketitle
\textbf{This version is an old draft, the final version of this article has been published in Frontiers In Neuroscience as:}\\
"Event-Based Gesture Recognition With Dynamic Background Suppression Using Smartphone Computational Capabilities"\\ \\
\textbf{and is freely available in open access at:}\\ \url{https://www.frontiersin.org/article/10.3389/fnins.2020.00275}
\\
\\
\textbf{The NavGesture dataset is available for download at:}\\
\url{https://www.neuromorphic-vision.com/public/downloads/navgesture/}
\\
\\
\textbf{Please cite as:}\\
@ARTICLE\{maro2020event,\\
AUTHOR=\{Maro, Jean-Matthieu and Ieng, Sio-Hoi and Benosman, Ryad\},\\
TITLE=\{Event-Based Gesture Recognition With Dynamic Background Suppression Using Smartphone Computational Capabilities\},\\
JOURNAL=\{Frontiers in Neuroscience\},\\
VOLUME=\{14\},\\
PAGES=\{275\},\\
YEAR=\{2020\},\\
URL=\{https://www.frontiersin.org/article/10.3389/fnins.2020.00275\},\\
DOI=\{10.3389/fnins.2020.00275\},\\
ISSN=\{1662-453X\},\\
\}
\afterpage{\blankpage}
  \end{@twocolumnfalse}
]

\clearpage\mbox{}\clearpage


\begin{abstract}
This paper introduces a framework of gesture recognition operating on the output of an event based camera using the computational resources of a mobile phone. We will introduce a new development around the concept of time-surfaces modified and adapted to run on the limited computational resources of a mobile platform. We also introduce a new method to remove dynamically backgrounds that makes full use of the high temporal resolution of event-based cameras. We assess the performances of the framework by operating on several dynamic scenarios in uncontrolled lighting conditions indoors and outdoors. We also introduce a new publicly available event-based dataset for gesture recognition selected through a clinical process to allow human-machine interactions for the visually-impaired and the elderly. We finally report comparisons with prior works that tackled event-based gesture recognition reporting comparable if not superior results if taking into account the limited computational and memory constraints of the used hardware.
\end{abstract}

\section{Introduction}
\begin{figure}[h!]
\centering
\includegraphics[width=0.9\columnwidth]{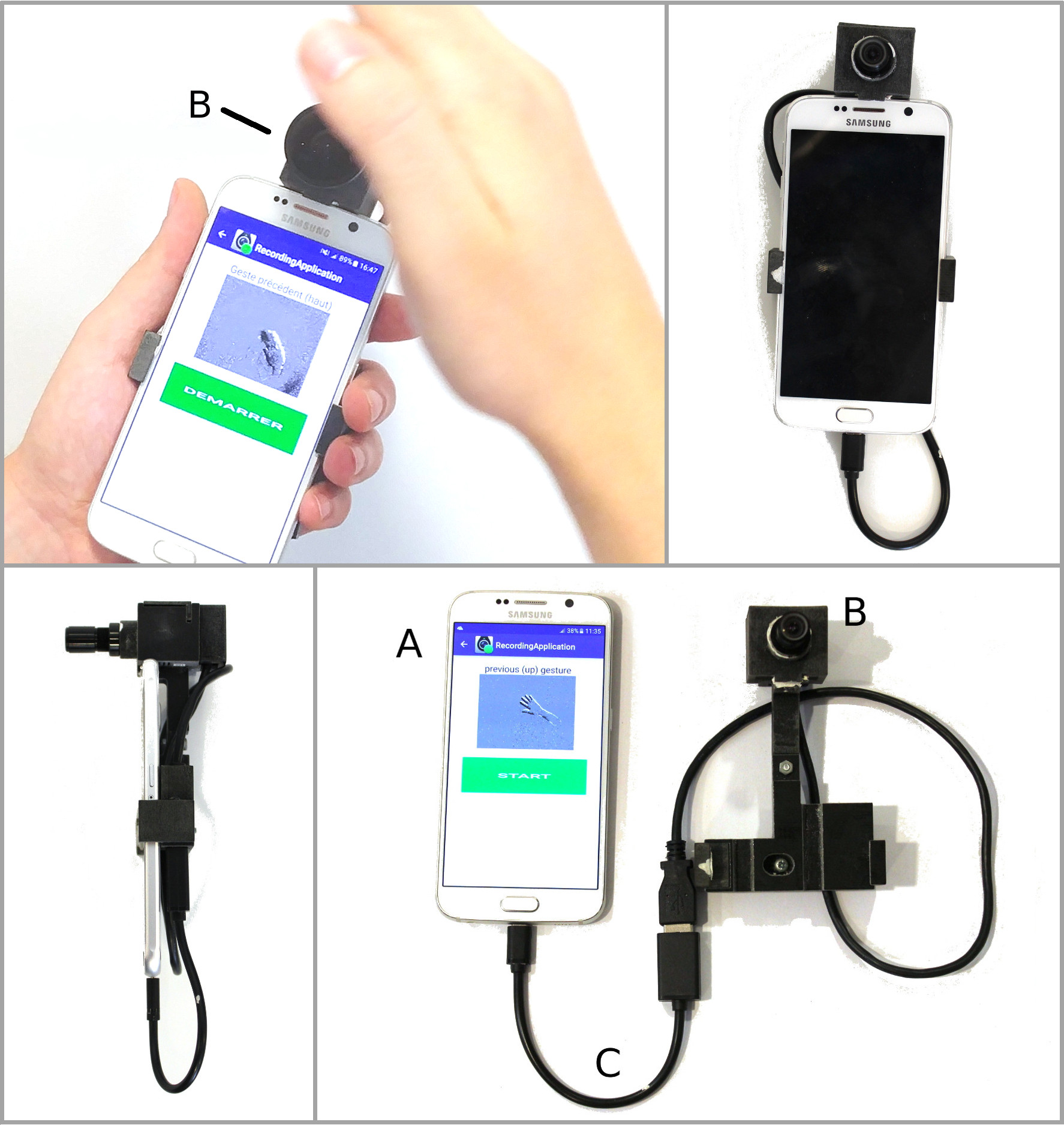} \\
\caption{A neuromorphic camera (an ATIS) (B) is plugged into a smart-phone (A) using an USB link (C), allowing mid-air gesture navigation on the smart-phone.}
\label{fig:phone_atis}
\end{figure}

This paper focuses on the problem of gesture recognition and dynamic background suppression using the output of a neuromorphic asynchronous camera \cite{Delbruck10,Posch14}. It allows for the first time to operate on the true dynamics of observed scenes event per event while only using the mobile phone computation capabilities without requiring connecting to off-board resources (Fig.\ref{fig:phone_atis}).

Neuromorphic event-based cameras offer a novel path to computer vision by allowing to operate at high temporal resolutions at equivalent frame rates at the order of several kHz at low computational power. They allow for a new level of performance in real-time vision with a drive towards more efficient algorithms.
Event-based cameras rely on a new principle that naturally allows all of the information contained in a standard video stream of several megabytes to be compressed in an event stream of a few kilobytes \cite{Lichtsteiner08,Posch11}.

In this paper we introduce a new method allowing for outdoor vision-based gesture recognition in real-time using the only computational power of a mobile phone. It features a scalable machine learning architecture relying on the concept of temporal surfaces introduced in \cite{lagorce2016hots} and extending it to operate more robustly. It also tackles the difficult problem of dynamic background suppression by introducing a novel approach in the temporal domain to this issue. Furthermore, we introduce a new data-set of gestures recorded using an event-based camera, that is made available publicly as the neuromorphic field still lacks data-sets that take full advantage of the precise timing of event-based cameras. Indeed, in most available data-sets such as N-MNIST and N-Caltech101 \cite{orchard2015converting} the dynamical properties of the data are artificially introduced. Even datasets such as the Poker Pips \cite{serrano2015poker} do not contain \textit{intrinsic} dynamical properties that could be used for classification. Compared to previous approaches, we emphasize the importance of using the information carried out by the timing of past events to obtain a robust low-level feature representation.
Driven by brain-like asynchronous event based computations, the methodology opens new perspectives for the Internet of Things (IOT) by operating asynchronously and allocating computational resources only on active parts of the network thus achieving lesser power consumption and faster response times. This differs from conventional image-based artificial neural network that require tremendous computational off-chip resources both for training and inference.

\section{Related Work}

Gesture recognition is an area of research that is quickly expanding \cite{pisharady2015recent, rautaray2015vision} and that currently relies mainly on two main streams of research. The first one uses wearable devices that mainly target specific indoors applications such as special effects, and are unsuited for outdoor use. The second stream relies on machine learning techniques coupled with several types of sensors. However, resource-constrained devices such as smart-phones disallow the use of certain technologies, such as vision-based depth sensors, due to their high energy consumption. This leads to the use of a wide variety of sensors, such as the proximity sensor \cite{cheng2011contactless, kim2010mobile} which is readily available on most smart-phones or even an off-board chip with radio-frequency capabilities as in \cite{kellogg2014bringing}. Considering vision-based approaches, several techniques have been developed to handle gesture recognition \cite{Mitra07} such as orientation  histogram \cite{Freeman95}, hidden Markov  models \cite{Starner97}, particle filtering \cite{Bretzner02}, support vector machine (SVM) \cite{Dardas11} and more recently convolution neural networks that allow featureless methodology \cite{DBLP:journals/corr/Xu17}. A vision-based method using only the built-in RGB camera of a smart-phone was introduced in \cite{song2014air}, but is limited to \textit{static} gestures (hand poses), excluding \textit{dynamic} gestures. The first gesture recognition system to take advantage of neuromorphic cameras to our knowledge was a stereo-vision setup, proposed by \cite{lee2014real}. In their work they use Leaky Integrate-and-Fire (LIF) neurons to correlate space-time events, in order to extract the trajectory of the gesture. Another work proposed a motion-based feature \cite{clady2016motion} that decays depending on the speed of the optical flow, which allows to take into account varying speeds. IBM research \cite{amir2017low} proposed an end-to-end neuromorphic system, running in real-time, using a Dynamic Vision Sensor (DVS) connected to a TrueNorth neuromorphic chip that performs the classification using a CNN-based architecture. Authors released a dataset, DvsGesture, which we use in our experiments, obtaining comparable results while being truly event-driven in the learning and inference.
This paper also goes beyond existing background suppression methodologies by using the high temporal resolution of event based cameras and thus allowing to operate on the activity of scenes rather than considering a frame based approach. This approach drastically contrasts from any existing background removal algorithm and does not rely on code-books \cite{Elgammal00}, probabilistic approaches \cite{Stauffer99}, sample-based methods \cite{Barnich11}, subspace-based techniques \cite{Oliver00} or even deep learning \cite{Babaee:2018:DCN:3178570.3178634}.

\section{Event-based Cameras and the Event-based Paradigm}

\begin{figure}[h]
\centering
\includegraphics[width=0.95\columnwidth]{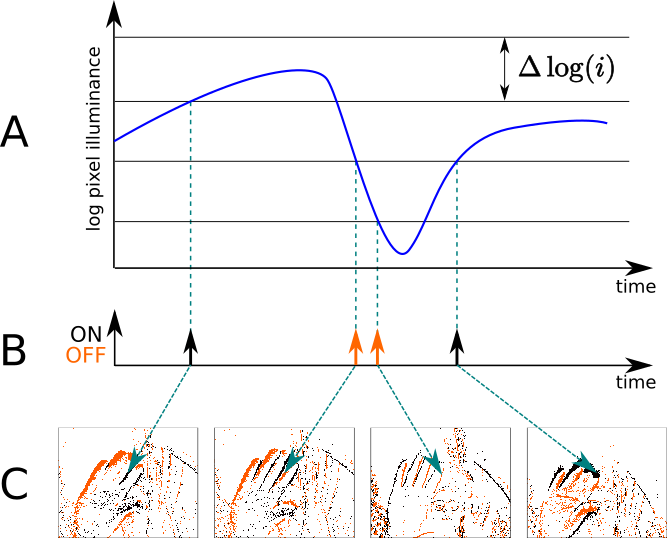} \\
\caption{Principle of operation of the neuromorphic camera used in this work. (A) When the change in illuminance of a given pixel's field of view exceeds a certain threshold, (B) it emits a visual event, which is either "ON" or "OFF" depending on the sign of the change. (C) A given pixel responds asynchronously to the visual stimuli in its own field of view.}
\label{fig:atisprinciple}
\end{figure}

The Address Event Representation (AER) neuromorphic camera used in this work is the Asynchronous Time-based Image Sensor (ATIS) (see Fig. \ref{fig:phone_atis}B) \cite{posch2011qvga}. Each pixel is fully autonomous, independent and asynchronous, and will only be triggered by a change in contrast in its own field of view. A pixel emits a visual event when the luminance change exceeds a certain threshold, typically 15\% percent in contrast. The nature of this change is encoded in the \textit{polarity} $p$ of the visual event, which can be either ON ($p = 1$) or OFF ($p = 0$), depending on the sign of the luminance change (see Fig. \ref{fig:atisprinciple}). The ATIS has a high temporal precision, in the order of the millisecond, which allows for the capture of highly dynamical scenes. Furthermore, static scenes will produce no output. This results in a low-redundancy, sparse and activity-driven stream of events at the output of such neuromorphic cameras.

The $k$-th visual event $e_k$ of the output stream of the camera can be mathematically written as the following triplet:
\begin{equation}
e_k = (\bm{x_k}, t_k, p_k)^\top
\end{equation}
where $\bm{x_k}$ is the spatial location of the visual event on the focal plane, $t_k$ its time-stamp, and $p_k$ its polarity.


\section{Method}

\subsection{Dynamic Background Suppression}

The Dynamic Background Suppression (DBS), aims to remove visual events that are not part of the useful signal, such as background objects. We make use of the native property of high temporal resolution of event-based cameras  that allows gestures to  generate a higher density of events than background objects as they closer to the camera.

\begin{figure}[h!]
\centering
\includegraphics[width=0.95\columnwidth]{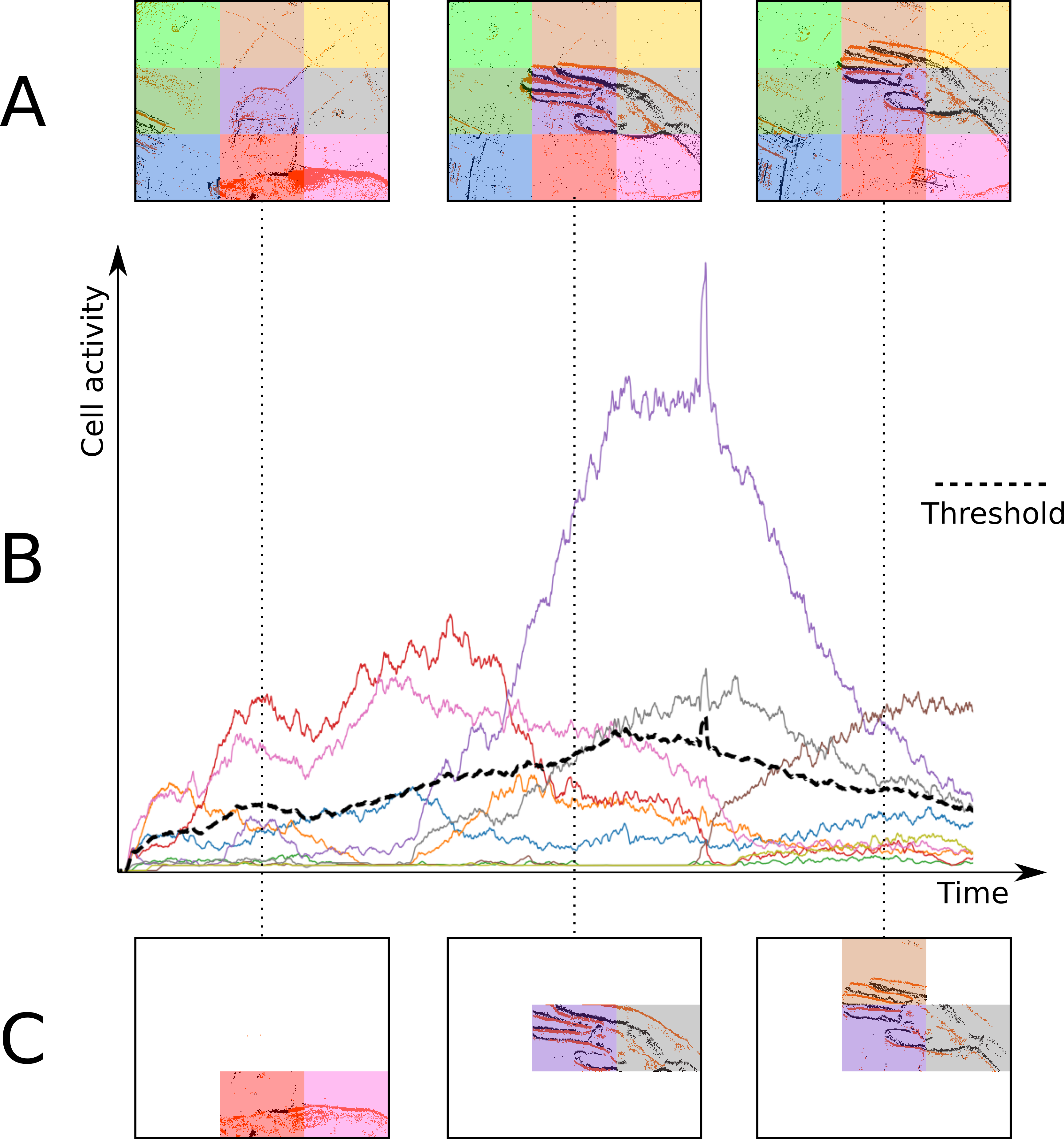}
\caption{Operating principle of the Dynamic Background Suppression (DBS). (A) A gesture is performed in front of the camera, which pixel array is divided into cells. (B) Each cell has its own activity counter that decays over time. (C) Only cells with their activity greater than the mean activity (black dashes) of all cells can spike.}
\label{fig:DBS}
\end{figure}

\begin{figure*}[h]
\centering
\includegraphics[width=1.9\columnwidth]{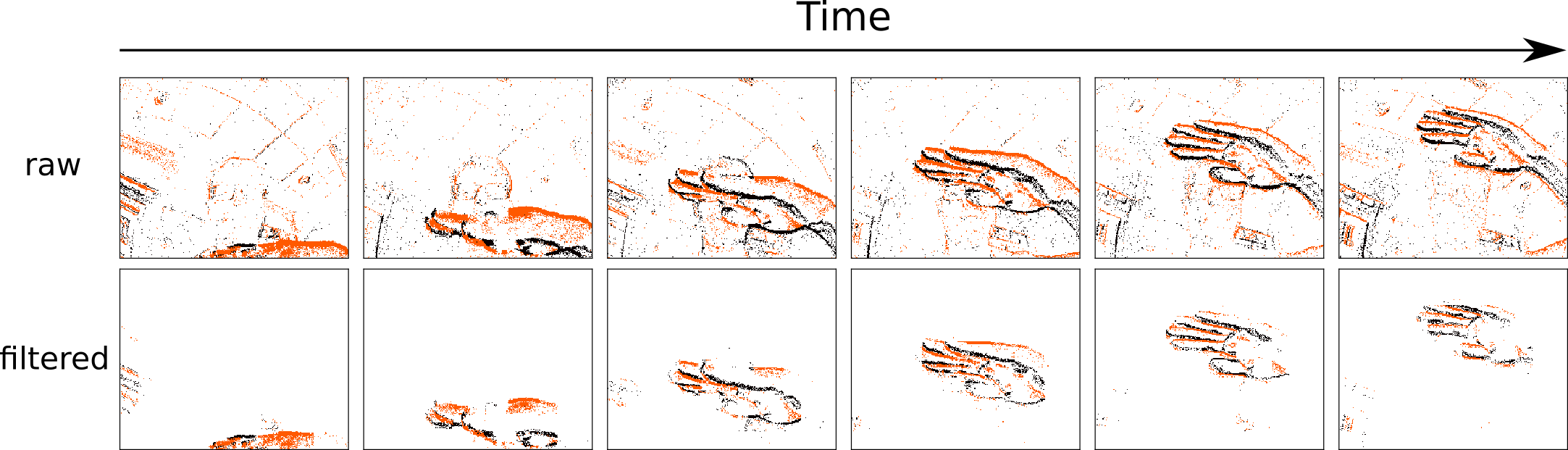} \\
\caption{Denoising example of a gesture clip from the NavGestures-walk data-set. The presented gesture is a "swipe down". Top row is the raw stream of visual events, and the bottom row is the denoised stream, at the output of the 3rd stage of the cascade presented in this paper. Each snapshot from the top row is made of 10,000 events, and bottom row contains only the kept events of those 10,000. "ON" events are orange, "OFF" events are black. The filtering lead to the removal of 83.8\% of all events. Even after removing this many events each gesture is still easily recognizable by the human eye.}
\label{fig:denoising}
\end{figure*}
The pixel array is divided into a grid of cells. Each cell $c$ hence contains several pixels and has its own activity counter noted $A_c$. At each event $e_k^c$ emitted by a pixel contained in cell $c$ we apply the following formula to update $A_c$:

\begin{equation}
A_c \leftarrow A_c \cdot \exp(- \frac{t_k^c - t_c}{ \tau_{b}}) + 1
\end{equation}

where $t_k^c$ is the time-stamp of the current event $e_k^c$, $t_c$ the last time a pixel spiked in the cell $c$, and $\tau_{b}$ a time-constant set in regards to the pixel array spike-rate. The average activity $\overline{A}$ of all cells is computed, namely only the events in cells with $A_c \geq \alpha\overline{A}$ (with $\alpha$ as a scalar to tune the aggressiveness of the filter) are propagated to the recognition module. This last stage prevents cells with a low spike-density, which are considered as background, from emitting events.
An example of the cascade operating on data from the NavGesture-walk dataset is shown in Fig. \ref{fig:denoising}.

\subsection{Time-surfaces as spatio-temporal descriptors}
A time-surface \cite{lagorce2016hots} is a descriptor of the spatio-temporal neighborhood of an event $e_k$.
We first define the time-context $T_k(\bm{u}, p)$ of the event $e_k$ as a map of time differences between the time-stamp of the current event and the time-stamps of the most recent events in its spatial neighborhood. This $(2R + 1) \times (2R + 1)$ map is centered on $e_k$, of spatial coordinates $\bm{x_k}$. The time-context can be expressed as:
\begin{equation}
T_k(\bm{u}, p) = \{ t_k - t \; | \; t = \max_{j \leq k} \, \{t_j \; | \; \bm{x_j} = (\bm{x_k} + \bm{u}), \; p_j = p\}\}
\end{equation}
where $\bm{u} = [u_x, u_y]^T$ is such that $u_x \in \llbracket-R, R\rrbracket$ and $u_y \in \llbracket-R, R\rrbracket$.

\begin{figure}[h!]
\centering
\includegraphics[width=0.95\columnwidth]{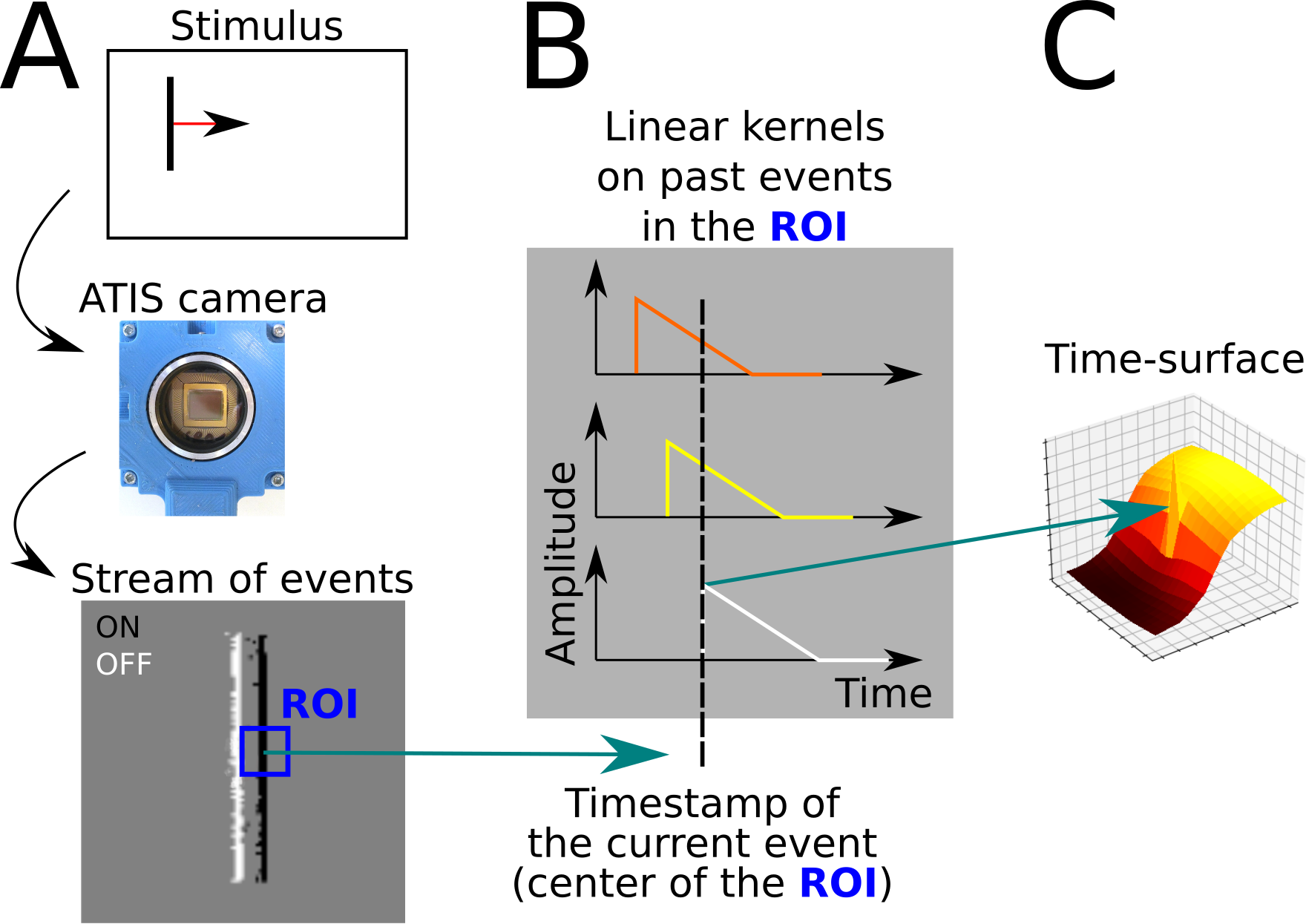}
\caption{(A) A moving vertical bar is presented to the event-based camera, which output a stream of visual events. The edges of the bar are ON (white) and OFF (black) events. A ROI is defined around the current event (blue square). (B) The time-stamps of visual events contained the ROI are decayed using a linear kernel. (C) The resulting extracted time-surface, that encodes both the contour orientation and the dynamic of the motion.}
\label{fig:ts}
\end{figure}

Finally, we obtain the time-surface $S_k(\bm{u}, p)$ associated with the event $e_k$, by applying a linear decay kernel of time-constant $\tau$ to the time-context $T_k$:
\begin{equation}
\label{eq:kernel}
    S_k(\bm{u}, p) =
\begin{cases}
    1 - \frac{T_k(\bm{u}, p)}{\tau}, & \text{if }  T_k(\bm{u}, p) < \tau\\
	0, 							     & \text{otherwise}
\end{cases}
\end{equation}
This gives a low-level representation of a local spatio-temporal neighborhood. However, as a time-surface is computed for each new incoming event, overlapping time surfaces are computed several times leading to resources wastes. In order to limit this effect, time-surfaces are discarded if they do not contain sufficient information, as this information will be part of a later time-surface as soon as a new event is emitted in the spatio-temporal neighborhood. For a time-surface to be considered valid, it must satisfy the following constraint:
\begin{equation}
\label{eq:valid}
\sum^{u, p}{S_k(\bm{u}, p)} \geq 2R
\end{equation}
where $R$ is the radius (half-width) of the time-surface. As the event-based camera performs native contour extraction, this ensures that sufficient events to form a valid descriptor can be carried out.

\subsection{Event-based hierarchical network}

The event-based camera visual events are fed to a network composed of several layers integrating information  over increasing temporal scales.
As information flows into the network, only their polarities "or new feature planes" are updated. Polarities in the network correspond to learned patterns or elementary features at that temporal and spatial scale. However, as events can be discarded, the network output stream usually contains less events than the input stream, which is an important property that build on the native low output of the event-based camera to lower the computational cost.

\subsubsection{Learning prototypes}
An iterative online clustering method is used to learn the patterns (hereinafter called prototypes), as it allows to process events as they are received, in an event-based manner. First, a set of $N$ time-surface prototypes $C_n$, with $n \in \llbracket0,N-1\rrbracket$, is created. The $C_n$ are initialized by simply using the first $N$ time-surfaces obtained from the stream of events. Then, for each incoming event $e_k$ we compute its associated time-surface $S_k$. Using the L2 Euclidean distance we compute the closest matching prototype $C_i$ in the bank, which we update with $S_k$ using the following rule:
\begin{equation}
\label{eq:update}
C_i \leftarrow C_i + \alpha_i \frac{S_k \cdot C_i}{\| S_k \| \; \| C_i \|} (S_k - C_i)
\end{equation}
with $\alpha_i$ the current learning rate of $C_i$ defined as: $$\alpha_i = \frac{1}{1 + A_i}$$
where $A_i$ is the number of time-surfaces which have already been assigned to $C_i$.
If a prototype $C_i$ has not been triggered by any of the last time-surfaces, it is initialized and forced to learn a new pattern.  This prevents badly initialized prototypes to stay unused, and helps them converge to meaningful representations while maintaining always on learning capabilities.
It is important to emphasize that compared to the original \cite{lagorce2016hots} we show that a linear decay (less computational expensive that the original exponential), combined with the heuristic that allows the suppression of a systematic computation of the time-surfaces allow for massive reduction in computation costs.
\subsubsection{Building the network}
\label{sec:net}
\begin{figure}[h]
\centering
\includegraphics[width=0.95\columnwidth]{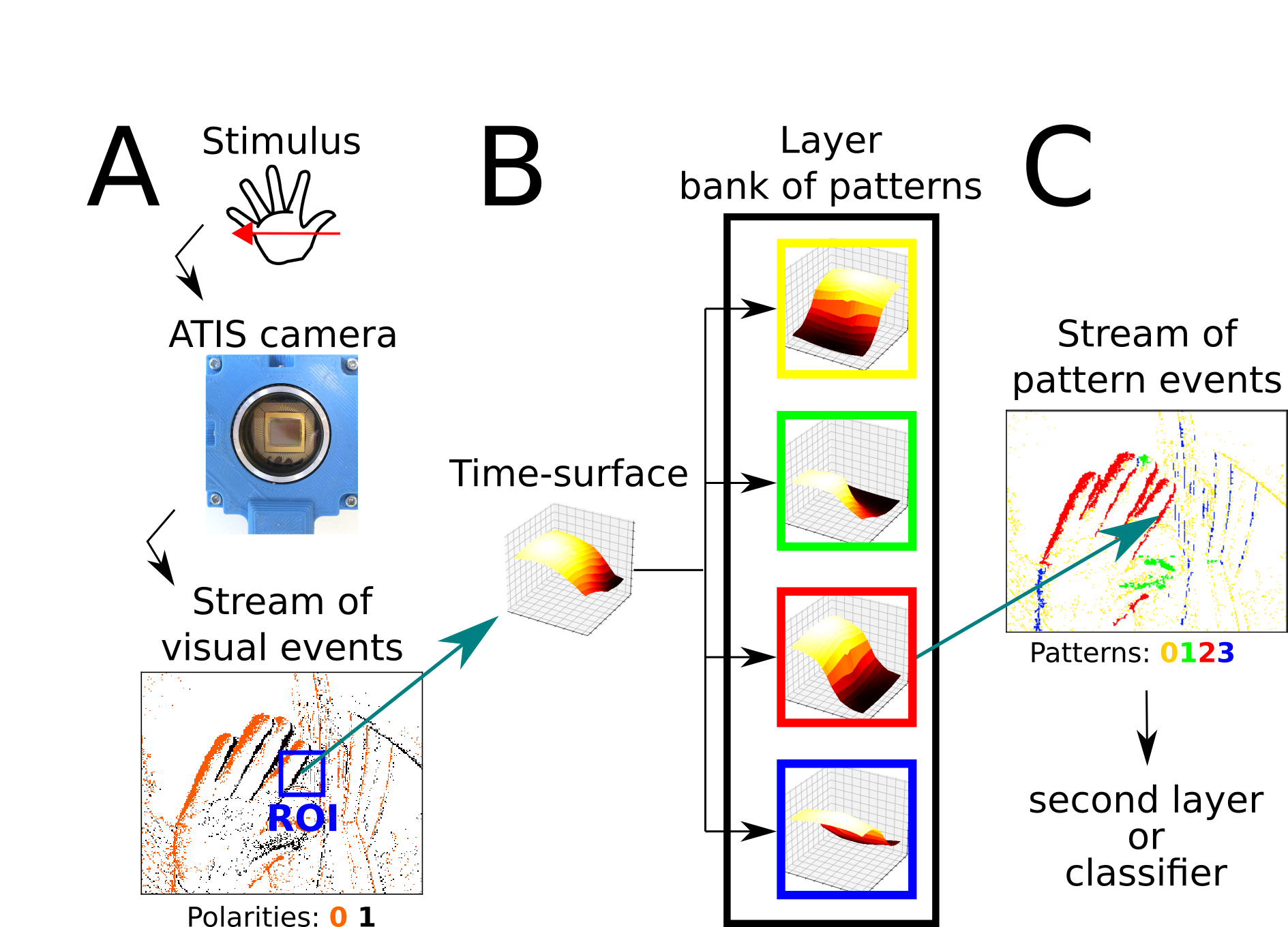}
\caption{(A) A stimulus is presented in front of a neuromorphic camera, which encodes it as a stream of event. (B) A time-surface can be extracted from this stream. (C) This time-surface is matched against known pattern, which are also time-surfaces, and that can be used as features for classification.}
\label{fig:hotsnetwork}
\end{figure}

The set of prototypes can be organized in a hierarchical manner (a set is then called a \textit{layer}), in order to form a network (see Fig. \ref{fig:hotsnetwork}). These layers can have different number of prototypes $N$, radius $R$ (which corresponds to a neuron's receptive field) and time-constant $\tau$.

The stimulus is presented to the event-based camera (Fig. \ref{fig:hotsnetwork}A), which outputs a stream of visual events. A given event $e_m$ of the stream must go through all the layers before the next one $e_{m+1}$ is processed. At each layer, the time-surface $S_m$ associated to $e_m$ is computed (see Fig. \ref{fig:hotsnetwork}B), using the previously introduced kernel in Eq. (\ref{eq:kernel}), of time constant $\tau$, and considering a spatial receptive field of side-length ($2R + 1$). If $S_m$ satisfies Eq. (\ref{eq:valid}), we update the closest prototype $C_c$ using Eq. (\ref{eq:update}) (see Fig. \ref{fig:hotsnetwork}B), and the polarity $p_m$ of $e_m$ is modified so that $p_m = c$, $c$ being the ID of the matching prototype. The polarity now encodes a pattern, and we talk of \textit{pattern events} instead of \textit{visual events} for which the polarity corresponds to a luminance change. The pattern event is then fed to the second layer, and processed in a similar manner. The second layer combines patterns from the first layer, thus its prototypes (and so the corresponding polarities) encode more sophisticated patterns. The second layer is therefore able to encode changes of direction in the motion. Once the full network has been trained, meaning that its time-surface prototypes have converged, the learning is disabled: prototypes are not updated using Eq. (\ref{eq:update}) anymore. The network can now serve as a feature extractor: the polarities of events output by the network will be used as features for classification.

\begin{table}[h!]
\centering
\begin{tabular}{|c|c|c|}
\hline
\textbf{Gest.}  & \textbf{\begin{tabular}[c]{@{}c@{}}Mean Event\\ Number\end{tabular}} & \textbf{\begin{tabular}[c]{@{}c@{}}Mean Percentage\\ left after DBS\end{tabular}} \\ \hline
\textbf{down}   & 988,901                                                              & 41\%                                                                              \\ \hline
\textbf{home}   & 2,398,850                                                            & 48\%                                                                              \\ \hline
\textbf{left}   & 969,014                                                              & 42\%                                                                              \\ \hline
\textbf{right}  & 962,501                                                              & 43\%                                                                              \\ \hline
\textbf{select} & 1,212,222                                                            & 30\%                                                                              \\ \hline
\textbf{up}     & 1,110,652                                                            & 44\%                                                                              \\ \hline
\end{tabular}
\caption{Mean percentage of events left after each the Dynamic Background Suppression for each gesture class.}
\label{table:denoising}
\end{table}

\begin{table*}[h!]
\centering
\begin{tabular}{|l|l|c|c|c|c|c|c|c|c|c|c|}
\hline
\multicolumn{1}{|c|}{}                            & \multicolumn{1}{c|}{}                                         &                                     &                        & \multicolumn{3}{c|}{Layer 1} & \multicolumn{3}{c|}{Layer 2} &                              &                           \\ \cline{5-10}
\multicolumn{1}{|c|}{\multirow{-2}{*}{ID}}        & \multicolumn{1}{c|}{\multirow{-2}{*}{Dataset}}                & \multirow{-2}{*}{DBS} & \multirow{-2}{*}{Grid} & N       & R      & Tau       & N       & R     & Tau        & \multirow{-2}{*}{Classifier} & \multirow{-2}{*}{Results} \\ \hline
E1                                                & Faces                                                         &                                     &                        & 32      & 6      & 5 ms      &         &       &            & k-NN, k = 1                  & 96.6\%                    \\ \hline

E2                                                & Faces                                                         &                                     &                        & 48      & 6      & 5 ms      &         &       &            & k-NN, k = 1                  & 97.9\%                    \\ \hline
E3                                                & Faces                                                         &                                     &                        & 64      & 6      & 5 ms      &         &       &            & k-NN, k = 1                  & 98.5\%                    \\ \hline

E4                                                & NavGestures-sit                                               & \checkmark                           &                        & 8       & 2      & 10 ms     &         &       &            & k-NN, k = 7                  & 94.5\%                    \\ \hline
E5                                                & NavGestures-sit                                               & \checkmark                           &                        & 8       & 2      & 10 ms     & 8       & 2     & 100 ms     & k-NN, k = 7                  & 95.9\%                    \\ \hline

E6                                                & NavGestures-walk                                              &                                     &                        & 8       & 2      & 10 ms     &         &       &            & k-NN, k = 7                  &          75.9\%                 \\ \hline
E7                                                & NavGestures-walk                                              &                                     &                        & 8       & 2      & 10 ms     & 8       & 2     & 100 ms     & k-NN, k = 7                  & 81.3\%                    \\ \hline

E8                                                & NavGestures-walk                                              & \checkmark                           &                        & 8       & 2      & 10 ms     &         &       &            & k-NN, k = 7                  & 88.7\%                    \\ \hline
E9                                                & NavGestures-walk                                              & \checkmark                           &                        & 8       & 2      & 10 ms     & 8       & 2     & 100 ms     & k-NN, k = 7                  & 92.6\%                    \\ \hline

E10                                               & DvsGestures 10cl                                              &                                     & 3$\times$3                & 8       & 2      & 10 ms     & 64      & 2     & 100 ms     & k-NN, k = 11                 & 96.6\%                    \\ \hline
E11                                               & DvsGestures 11cl                                              &                                     & 3$\times$3                & 8       & 2      & 10 ms     & 64      & 2     & 100 ms     & k-NN, k = 11                 & 88.9\%                    \\ \hline

E12 & DvsGestures 11cl &                                     & 5$\times$5                & 8       & 2      & 10 ms     & 64      & 2     & 100 ms     & MLP                          & 93.1\%                    \\ \hline
\end{tabular}
\caption{Detail of the experiments that were taken on the different datasets.}
\end{table*}


\section{Datasets}
We used four datasets, all of them were recorded using a neuromorphic camera: Faces dataset \cite{lagorce2016hots}, DvsGesture \cite{amir2017low} and two novel datasets, NavGestures-sit and NavGestures-walk, tailored to facilitate the use of a smartphone by the elderly and the visually-impaired. NavGestures datasets are publicly available at [url hidden during reviewing for anonymity purposes].

\subsection{NavGestures-sit and NavGestures-walk Datasets}
The NavGestures-sit dataset was designed to operate on a smartphone using mid-air gestures. The gesture dictionary has only 6 gestures in order to be easily memorable but have also been determined as being the most elementary and sufficient set to operate a mobile phone. Four of them are "sweeping" gestures: \textit{Right, Left, Up, Down}. These are designed to navigate through the items in a menu. The \textit{Home} gesture, a "hello"-waving hand, can be used to go back to the main menu, or to obtain help. Lastly, the \textit{select} gesture, executed only using fingers, closing them as a claw in front of the device, and then reopening them, is used to select an item.

The dataset features 35 subjects, 12 being visually-impaired subjects, with a condition ranging from 1 to 4/5 on the WHO blindness scale and 23 being people from the laboratory. The gestures were recorded in real use condition, with the subject sitting and holding the phone in one hand while performing the gesture with their other hand. Some of the subjects were shown video-clips of the gestures to perform, while some others had only an audio description of the gesture. This inferred some very noticeable differences in the way each subject performed the proposed gestures, in terms of hand shape, trajectory, motion and angle but also in terms of the camera pose. Each subject performed 10 repetitions of the 6 gestures. Then all the gesture clips were manually labelled and segmented. We removed clips with a wrong field of view, wrongly executed gestures or that had a device-related capturing issues. The manually curated dataset contains $1, 621$ clips.

The NavGestures-walk dataset contains the 6 same gestures. The main difference being that the users walked through an urban environment while holding the phone with one hand and performed the gestures with the other. The dataset features 10 people from the laboratory that performed several times each of the 6 gestures. The dataset was recorded in uncontrolled lighting condition, both indoor in the laboratory, and outdoor in the nearby streets.
\subsection{DvsGesture and Faces Dataset}
IBM Research released a 10-class (plus a rejection class with random gestures) dataset \cite{amir2017low} of hand and arm gestures, performed by 29 subjects under 3 different lighting conditions. The camera is mounted on a stand, and the subjects stand still in front of it, therefore the database is lacking dynamic backgrounds at the core of our work but provides valuable grounds for comparisons. Authors split the dataset into a train database consisting of 23 subjects and a test database consisting of the 6 remaining subjects.\\
The Faces dataset \cite{lagorce2016hots}, contains clips of the faces of 7 subjects recorded using an event-based camera. Each subject made 24 recordings, resulting in 168 clips. The subjects moved their face while following a dot on a computer screen in a square movement. The dynamic is therefore the same for all subjects, and does not carry any meaningful information for the classification task. The faces dataset does not come with a proposed split between a train and a test subset. This allows us to perform cross-validation (10 random shuffles of train and test subsets) to ensure that the results are solid. As in the original paper, we put 5 examples in the train subset, and 19 the test subset.


\section{Experiments and Results}

In the following experiments we did not take the polarity of visual events into account: we considered that only the illuminance \textit{change} carries information for these classification tasks, and not the fact that the illuminance \textit{increased} or \textit{decreased}. This is because the same gesture can generate either ON or OFF events depending on the skin color, the clothing color or the background.
For all classification tasks, the output of end-layers is integrated over time to generate a histogram of activity per feature. This can then be used as a dynamic signature of the observed stimulus that can then be fed to a classifier (here a nearest neighbor). More sophisticated classifiers can be used, we chose however partly to save power resources a simple methodology but mostly to show that extracted features are strongly capturing the essence of the dynamic signature of the recorded gestures.

\subsection{Removing the background on the NavGestures datasets}
If subjects are sitting in the NavGestures-sit, they do hold the phone in their hand, which results in movements and unwanted jitters that both generate  background activity. In the case of the NavGestures-walk the visual background is even more present as subjects were walking while recording the dataset. Figure \ref{fig:denoising} illustrates the use of the Dynamic Background Suppression (DBS). Table \ref{table:denoising} reports the mean percentage of events left for each gesture class after removing the background. It must be noted that we did not use the DBS on the DvsGesture dataset because it was recorded with a static camera and because the background was static, therefore there is no background to remove.
The following parameters were used for the DBS:
\begin{itemize}
    \item $\tau_b = 300 \mu$s
    \item $\alpha = 2$
    \item grid size : $3 \times 3$
\end{itemize}



\subsection{Results on the gesture datasets}
In our experiments on the gesture datasets (E4 to E11) we tried both 1-layer and 2-layers networks, and also the benefits of the Dynamic Background Suppression on the recognition rate.
Two-layers network perform better, as they can handle changes in direction. Also the Dynamic Background Suppression greatly improves the recognition rate, as demonstrated for the NavGestures-walk, increasing the score from 81.3\% to 92.6\%.

Regarding the DvsGesture dataset, we use the same 2-layer network architecture. The only difference is that we increased the number of prototypes in the second layer because the gestures are more complex, and more prototypes in the end-layer account for more discriminative power.

We also took into account the spatial component of gestures. This is possible because clips of the DvsGesture dataset all have the same framing. We split the pixel array into sub-regions, using a $3 \times 3$ grid. Hence, the final feature is a histogram of size $3 \times 3 \times 64 = 576$. Classification used a nearest neighbor classifier on the histograms. One can observe in table \ref{table:comparisonresults} that the system performs in the same range of precision as \cite{amir2017low} while being lighter to implement and compute.

\begin{table}[]
\begin{tabular}{|l|c|c|c|}
\hline
\textbf{}                 & This work       & HOTS \cite{lagorce2016hots} & IBM \cite{amir2017low}    \\ \hline
Faces            & \textbf{98.5\%}   & 79\% & -      \\ \hline
DvsGestures 10 cl      & \textbf{96.6\%} & -    & 96.5\% \\ \hline
DvsGestures 11 cl      & 93.1\% & -    & \textbf{94.6\%} \\ \hline
NavGestures-sit  & \textbf{95.9\%}           & -    & -      \\ \hline
NavGestures-walk & \textbf{92.6\%}           & -    & -      \\ \hline
\end{tabular}
\caption{Comparison of the classification accuracy on event-based datasets.}
\label{table:comparisonresults}
\end{table}

\subsection{Gesture recognition on the smartphone}
\begin{figure}[h]
\centering
\includegraphics[width=0.95\columnwidth]{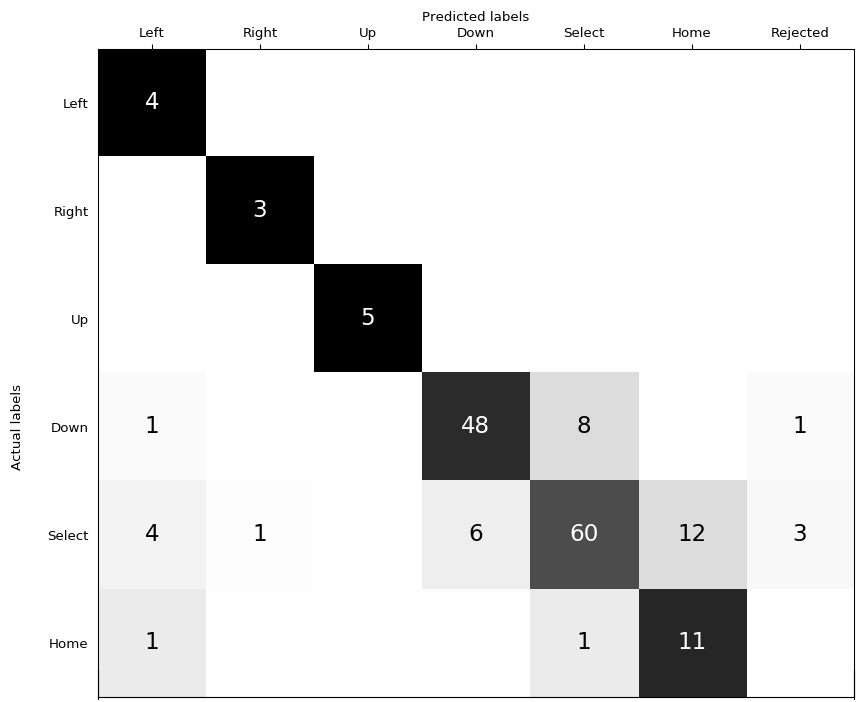}
\caption{Results obtained with visually-impaired end-users on early tests using the prototype. Users were asked to perform different scenarii such as sending a pre-written message. Average correct classification is 78\%, however this score is heavily impacted by not-so-good score of the "select" gesture, which is performed in very different ways among users, something we already had reckoned during the dataset creation.}
\label{fig:phoneresults}
\end{figure}

The whole system, made of the DBS, a 1-layer feature extractor and the recognition module, is implemented on a mobile phone, a Samsung GM-920F, as native C++ code. The event-based camera is directly plugged into the micro-usb port of the mobile phone (see Fig. \ref{fig:phone_atis}). This prototype was briefly tested by visually-impaired end-users, in real use condition. They were asked to perform certain tasks using the phone, such as sending a pre-written message or play a song. Results of the pre-tests can be found in table \ref{fig:phoneresults}. It is important to emphasize that some gestures require longer execution time, because it they generate much more visual event that thus require more computation. This is one of the properties of being scene-driven.

\subsection{Results on the Faces dataset}

Using a single-layer network with a receptive field R = 6, N = 32 prototypes and Tau = 5 ms, the scope is to push the system to it limit and inquire whether a single layer is enough to capture the static properties of this dataset. We are able to obtain 96.6\% recognition score on this dataset, whereas the original model in \cite{lagorce2016hots} performed at 79\% using a three-layer architecture, with its end-layer having the same number N = 32 of prototypes. When increasing the number of prototypes to N = 48 and N = 64, we achieved respectively 97.9\% and 98.5\% in average recognition rate. Also we noticed that increasing Tau higher than 5 ms was not beneficial and decreased our classification accuracy. The data properties in this dataset are static: the dynamic does not carry any meaningful information for classifying the faces. This shows that the numerous modifications we introduced into the model lead to an important improvement in extracting the static properties.

Additional material provides videos of the Dynamic Background Suppression at work and of live gesture recognition on the smartphone.


\section{Discussion}
In this work, we presented a system that allows to recognize gestures using a smartphone computational capabilities. We also improved drastically the hierarchical network proposed in \cite{lagorce2016hots}, both for static and dynamic data. The system and methodology allows to truly understand what the network is computing rather than the conventional black-box approach. We can report that the first layers operating on shorter time scales are extracting oriented contours and direction, while the second layers encode change of directions of the same feature. Deeper networks could theoretically encode multiple changes in direction, but given the nature of gestures and the task to be performed there no need to use such networks. We can assess that a 2-layered network is sufficient to handle efficiently any of the considered databases.
This is truly the advantage of using time-surfaces that encode in a compact representation both spatial and temporal information.
The system also relies on a very small number of meta parameters to tune. We did not require long parameter adjusting processes for all the considered databases. Once the parameters of the network match those of the observed object, the same set can apply regardless to the dataset, and we were able to use the same parameters for all the gestures datasets, while obtaining state-of-the-art accuracy scores on the DvsGestures classification task.
We believe this is the first time that time-surfaces were used at their true potential. Indeed, in previous work like HOTS \cite{lagorce2016hots} or HATS \cite{sironi2018hats} the decay times used were set to values thousands times higher than the duration of the stimulus. This resulted in time-surfaces that acted as binary frames, instead of truly encoding the dynamic of the scene. This comes as no surprise as considering inadequate time scales uncorrelated with the dynamics of the observed scene will provide low amounts of information and therefore poor recognition rates.
Finally, the Dynamic Background Suppression plays a very important role in achieving high recognition rates in a walking situation.

\section{Acknowledgments}
The authors would like to thanks Christopher Reeves for his tremendous help in the creation of the NavGestures-sit dataset. We also would like to thanks Antonio Fernandez, Andrew Watkinson and Gregor Lenz for their contribution in the android application.

\bibliographystyle{plain}

\end{document}